\documentclass[11pt,letterpaper]{article}
\usepackage{emnlp2016}
\usepackage{times}
\usepackage{latexsym}
\usepackage{amsmath}
\usepackage{graphicx}
\usepackage{multirow}
\usepackage{booktabs}
\usepackage{epstopdf}
\usepackage{caption}
\usepackage{amsfonts}
\usepackage{subcaption}
\emnlpfinalcopy

\title{Sharing Network Parameters for Crosslingual Named Entity Recognition}

\author{Rudra Murthy V \\ Indian Institute Of Technology \\ Bombay  \\ {\tt rudra@cse.iitb.ac.in}
	\And Mitesh Khapra \\ IBM Research India \\ {\tt mikhapra@in.ibm.com} 
	\And Dr. Pushpak Bhattacharyya \\ Indian Institute Of Technology \\ Bombay \\ {\tt pb@cse.iitb.ac.in}}

\date{}

\begin{document}

\maketitle

\begin{abstract}
Most state of the art approaches for Named Entity Recognition rely on hand crafted features and annotated corpora. Recently Neural network based models have been proposed which do not require handcrafted features but still require annotated corpora. However, such annotated corpora may not be available for many languages. In this paper, we propose a neural network based model which allows sharing the decoder as well as word and character level parameters between two languages thereby allowing a resource fortunate language to aid a resource deprived language. Specifically, we focus on the case when limited annotated corpora is available in one language ($L_1$) and abundant annotated corpora is available in another language ($L_2$). Sharing the network architecture and parameters between $L_1$ and $L_2$ leads to improved performance in $L_1$. Further, our approach does not require any hand crafted features but instead directly learns meaningful feature representations from the training data itself. We experiment with 4 language pairs and show that indeed in a resource constrained setup (lesser annotated corpora), a model jointly trained with data from another language performs better than a model trained only on the limited corpora in one language.        

\end{abstract}

\section{Introduction}
Named Entity Recognition (NER) plays a crucial role in several downstream applications such as Information Extraction, Question Answering, Machine Translation \textit{etc.}. Existing state of the art systems for NER are typically supervised systems which require sufficient annotated corpora for training  \cite{Ando:2005:FLP:1046920.1194905,Collobert:2011:NLP:1953048.2078186,Turian:2010:WRS:1858681.1858721}. In addition, they rely on language-specific handcrafted features (such as capitalization of first character in English). Some of these features rely on knowledge resources in the form of gazetteers \cite{Florian:2003:NER:1119176.1119201} and other NLP tools such as POS taggers which in turn require their own training data. This requirement of resources in the form of training data, gazetteers, tools, feature engineering, etc. makes it hard to apply these approaches to resource deprived languages.

Recently, several Neural Network based approaches for NER have been proposed \cite{Collobert:2011:NLP:1953048.2078186,DBLP:journals/corr/HuangXY15,737,yang2016multi,DBLP:journals/corr/GillickBVS15} which circumvent the need for hand-crafted features and thereby the need for gazetteers, part-of-speech taggers, etc. They directly learn meaningful feature representations from the training data itself and can also benefit from large amounts of unannotated corpora in the language. However, they still require sufficient data for training the network and thus only partially address the problem of resource scarcity.   

Very recently \newcite{DBLP:journals/corr/GillickBVS15} proposed an encoder decoder based model for sequence labeling which takes a sequence of bytes (characters) as input instead of words and outputs spans as well as labels for these spans. For example, in the case of part-of-speech tagging the span could identify one word and the associated label would be the part-of-speech tag of that word. Since the input consists of character sequences, the network can be jointly trained using annotated corpora from multiple languages by sharing the vocabulary (characters, in this case) and associated parameters. They show that such a jointly trained model can perform better than the same model trained on monolingual data. However, they do not focus on the resource constrained setup where one of the languages has very little annotated corpora. Further, the best results in their joint training setup are poor when compared even to the monolingual results reported in this paper.

In this paper, we propose a neural network based model which allows sharing of character dependent, word dependent and output dependent parameters. Specifically, given a sequence of words, we employ LSTMs at word level and CNNs at character level to extract complementary feature representations. The word level LSTMs can capture contextual information and the character level CNNs can encode morphological information. At the output layer we use a feedforward network to predict NER tags. Similar to \newcite{DBLP:journals/corr/GillickBVS15}, our character dependent parameters are shared across languages (which use the same character set). However, unlike \newcite{DBLP:journals/corr/GillickBVS15} we do not use an encoder decoder architecture. Further, our model also employs word level features which can be shared across languages by using jointly learned bilingual word embeddings from parallel corpora \cite{icml2015_gouws15}. Since the NER tags are same across languages, even the output layer of our model is shared across languages.  

We experiment with 4 language pairs, \textit{viz.}, English-Spanish, English-German, Spanish-German and Dutch-German using standard NER datasets released as part of the CoNLL shared task \cite{TjongKimSang:2003:ICS:1119176.1119195,TjongKimSang:2002:ICS:1118853.1118877} and German NER data by \newcite{faruqui10:_training}. We artificially constrain the amount of training data available in one language and show that the network can still benefit from abundant annotated corpora in another language by jointly learning the shared parameters. 
Further, in the monolingual setup we report state of the art results for two out of three languages without using any handcrafted features or gazetteers.

The remainder of this paper is organized as follows: 

\section{Related Work}
In this section we present a quick overview of (i) neural network based approaches for NER which now report state of the art results and (ii) approaches catering to multilingual NER.

Neural networks were first explored in the context of named entity recognition by \newcite{Hammerton:2003:NER:1119176.1119202} but, \newcite{Collobert:2011:NLP:1953048.2078186} were the first to successfully use neural networks for several NLP tasks including NER. Unlike existing supervised systems, they used minimal handcrafted features and instead relied on automatically learning word representations from large unannotated corpora. The output layer was a CRF layer which modeled the entire sequence likelihood. They also used the idea of sharing network parameters across different tasks  (but not between different languages).

This idea was further developed by \cite{icml2014c2_santos14,DBLP:journals:corr:SantosG15} to include character level information in addition to word level information. They used Convolutional Neural Networks (CNNs) with fixed filter width to extract relevant character level information. The combined character features and word embeddings were fed to a time delay neural network as in \newcite{Collobert:2011:NLP:1953048.2078186} and used for Spanish and Portuguese NER.

There are a few works which use Bidirectional Long Short Term Memory (Bi-LSTMs) \cite{650093}
for encoding word sequence information for sequence tagging. For examples \newcite{DBLP:journals/corr/HuangXY15} use LSTMs for encoding word sequences and then use CRFs for decoding tag sequences.  \newcite{DBLP:journals/corr/ChiuN15} use a combination of Bi-LSTMs with CNNs for NER. The decoder is still a CRF which was trained to maximize the entire sequence likelihood. Both these approaches also use some handcrafted features. Very recently \newcite{737} proposed Hierarchical Bi-LSTMs as an alternative to CNN-Bi-LSTMs wherein they first use a character level Bi-LSTMs followed by a word level Bi-LSTMs, thus forming a hierarchy of LSTMs. They also used CRF at the output layer. The model was tested on English, Spanish, Dutch, and German languages. They reported state-of-the-art results when systems with no handcrafted feature engineering are considered.

Very recently \newcite{DBLP:journals/corr/GillickBVS15} proposed a novel encoder-decoder architecture for language independent sequence tagging. Even more recently, \newcite{yang2016multi} extended \newcite{737} and focused on both multi-task and multilingual setting. In the multi-task scenario, except for the output CRF layer, the rest of the network parameters were shared. In the multilingual setting only the character-level features were shared across languages. Though they reported some improvements in the multilingual setting, their model is not suitable in a resource constrained setup (limited training data) because knowledge sharing between languages happens only through character-level features.

Multilingual training of NER systems was explored dating back to \cite{Babych:2003:IMT:1609822.1609823}. Usually these systems train  a language dependent NER tagger by (i) enforcing tag constraints along the aligned words in parallel tagged corpora \cite{Chen:2010:JRA:1858681.1858746,Li:2012:JBN:2396761.2398506} or untagged parallel corpus \cite{DBLP:conf/aaai/WangCM13,DBLP:journals/tacl/WangM14,DBLP:journals/tacl/WangM14,Wang_jointword} and/or (ii) use cross-lingual features  \cite{Li:2012:JBN:2396761.2398506,Tackstrom:2012:CWC:2382029.2382096,Che_namedentity}. 

Unlike existing methods, our proposed deep learning model allows sharing of different parameters across languages and can be jointly trained without the need for any annotated parallel corpus or any handcrafted features. 
\section{Model}
In this section, we describe our model which encodes both character level as well as word level information for Named Entity Recognition. As shown in Figure \ref{MonoNERPara}, our model consists of three components, {viz.}, (i) a convolutional layer for extracting character-level features, (ii) a bi-directional LSTM for encoding input word sequences and (iii) a feedforward output layer for predicting the tags. 

\subsection{Character level Convolutional Layer}
The input to our model is a sequence of words $X = \{x_1, \ldots, x_n\}$. We consider each word $w_i$ to be further composed of a sequence of characters, \textit{i.e.}, $x_i = \{c_{i1}, c_{i2}, ..., c_{ik} \}$ where $k$ is the number of characters in the word. Each character $c_{i1}$ is represented as a one hot vector $\in {R}^{|C|}$ where $|C|$ is the number of characters in the language. These one-hot representations of all the characters in the word are stacked to a form a matrix $\mathbf{M} \in {R}^{k \times |C|}$. We then apply several filters of \textit{one dimensional} convolution to this matrix.  The width of these filters varies from 1 to $n$, \textit{i.e.}, these filters look at 1 to $n$-gram character sequences. The intuition is that a filter of length 1 could look at unigram characters and hopefully learn to distinguish between upper case and lowercase characters. Similarly, a filter of length 4 could learn that a sequence "son\$" at the end of a word indicates a PERSON (as in Thomson, Johnson, Jefferson, etc). 

\begin{figure}
\centering
\includegraphics[width=0.5\textwidth]{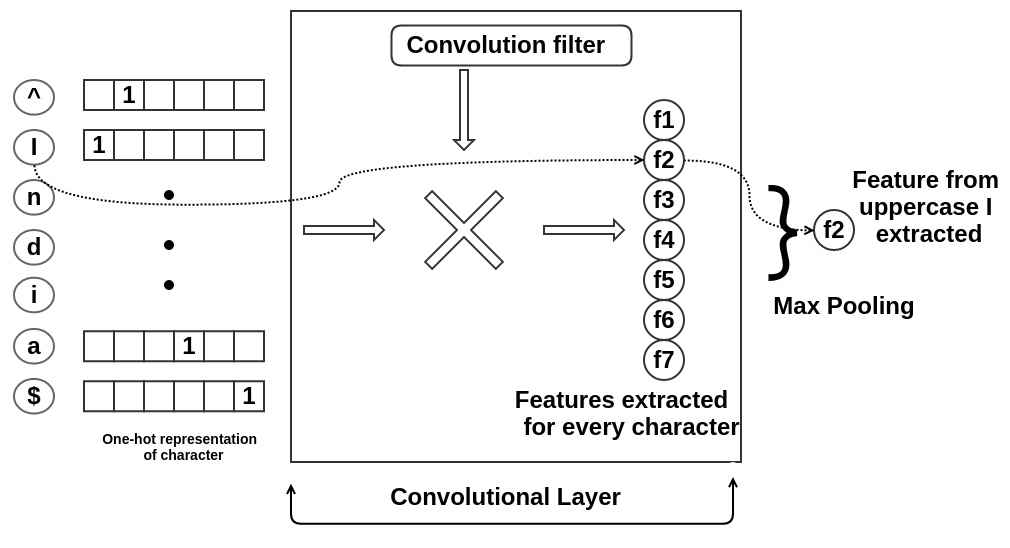}
\caption{Convolutional Neural Network extracting unigram features}
\label{fig:CNNfeature}
\end{figure}

The convolutional operation is followed by a max-pooling operation to pick the most relevant feature (for example, as shown in Figure \ref{fig:CNNfeature}, the max-pooling layer picks up the feature corresponding to capitalization). Further, since there could be multiple relevant n-grams of the same length we define multiple filters of each width. For example, each of the 4-gram sequences \textit{son\$}, \textit{corp}, \textit{ltd.} is relevant for NER and different filters of width 4 could capture the information encoded in these different 4-gram sequences. In other words, we have $k_1, k_2, ..., k_n$ filters on width 1,2, ..., n. If we have a total of $d_1$ such filters ($d_1 = k_1 + k_2 + ... + k_n$) then we get a $d_1$ dimensional representation of the word denoted by $h_{cnn}(x_i)$.   

\subsection{Bi-directional LSTM}
The input to the bi-directional LSTM is a sequence of words where each word is represented by the following concatenated vector. 

\begin{equation}
h(x_i) = [h_{emb}(x_i), h_{cnn}(x_i)]
\end{equation}

$h_{emb}(x_i)$ is simply the embedding of the word which can be pre-trained (say, using word2vec \cite{mikolov2013distributed,mikolov2013linguistic}) and then fine tuned while training our model. The second part, \textit{i.e.}, $h_{cnn}(x_i)$ encodes character level information as described in the previous sub-section. 

The forward LSTM reads this sequence of word representations from left to right whereas the backward LSTM does the same from right to left. This results in a hidden representation for each word which contains two parts. 

\begin{equation}
g_i = [f_i(x_1,\ldots,x_i), b_i(x_n,\ldots,x_i)]
\label{eq2}
\end{equation}

where, $f_i$ and $b_i$ are the forward and backward LSTM's outputs respectively at time-step (position) $i$. We use the standard definitions of the LSTM functions $f_i$ and $b_i$ as described in \newcite{DBLP:journals/corr/GillickBVS15}. 

\subsection{Decoder}

Given a training set $D = (X,Y)$ where $X = (x_1, \ldots, x_n)$ is a sequence of words and $Y = (y_1, \ldots, y_n)$ is a corresponding sequence of entity tags, our goal is to maximize the log-likelihood of the training data as in equation \ref{eq0}.
\begin{equation}
\underset{\theta}{\text{maximize }}  \sum_{\forall (X,Y) \in D} \log P(Y| X) 
\label{eq0}
\end{equation}
where $\theta$ are the parameters of the network. The log conditional probability $P(Y|X)$ can be decomposed as in equation \ref{eq1},
\begin{equation}
\log P(Y| X) = \sum_{i=1}^{n} \log P(y_i | x_1,\ldots,x_n, y_{i-1})
\label{eq1}
\end{equation}

We model $\log P(y_i | x_1,\ldots,x_n, y_{i-1})$ using the following equation:
\begin{multline}
\log P(y_i | x_1,\ldots,x_n, y_{i-1})  = W_{y_i}g_i + A_{y_iy_{i-1}}  - \\ \log \sum_{k\in T} \exp (W_{y_k}g_i + A_{y_ky_{i-1}}) 
\end{multline}

where, $W_{y_i}$ is a parameter vector w.r.t tag $y_i$ which when multiplied with $g_i$ gives a score for assigning the tag $y_i$. Matrix $A$ can be viewed as a transition matrix where the entry $A_{y_iy_{i-1}}$ gives the transition score from tag $y_{i-1}$ to tag $y_i$. $T$ is the set of all possible output tags. 

In simple words, our decoder computes the probabilities of the entity tags by passing the output representations computed by LSTM at each position $i$ and the previous tag $y_{i-1}$ through a linear layer followed by a softmax layer. In this sense, our model is a complete neural network based solution as opposed to existing models which use CRFs at the output.

\begin{figure}[!htb]
\centering
\includegraphics[width=6.5cm,height=8cm]{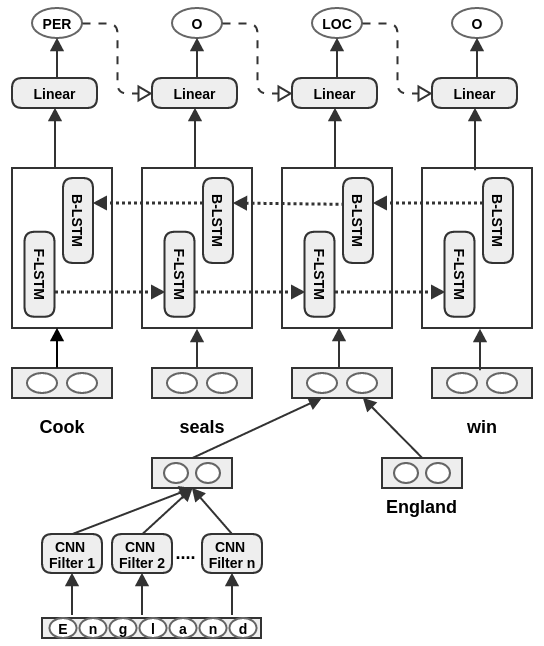}
\caption{Architecture of Proposed NER System}
\label{MonoNER}
\end{figure}

\subsection{Sharing parameters across languages}
\begin{figure}[!htb]
\centering
\includegraphics[width=5cm]{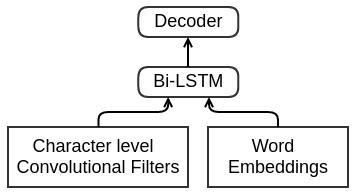}
\caption{Different Parameters in our NER System}
\label{MonoNERPara}
\end{figure}

As shown in Figure \ref{MonoNERPara}, our model contains the following parameters: (i) convolutional filters (ii) word embeddings (iii) LSTM parameters and (iv) decoder parameters. The convolutional filters operate on character sequences and hence can be shared between languages which share a common character set. This is true for many European languages and we consider some of these languages for our experiments (English, Spanish, Dutch and German). Recently there has been a lot of interest in jointly learning bilingual word representations. The aim here is to project words across languages into a common space such that similar words across languages lie very close to each other in this space. In this paper, we experiment with Bilbowa bilingual word embeddings which allows us to share the space of word embeddings across languages. Similarly, we also share the output layer across languages since all languages have the same entity tagset. Finally, we also share the LSTM parameters across languages. Thus, irrespective of whether the model sees a Spanish training instance or an English training, the same set of filters, LSTM parameters and output parameters get updated based on the loss function (and of course the word embeddings corresponding to the words present in the sentence also get updated).

\section{Experimental Setup}
In this section we describe the following: (i) the datasets used for our experiments (ii) publicly available word embeddings used for different languages and (iii) the hyperparameters considered for all our experiments. 

\subsection{Dataset}
For English, Spanish and Dutch we use the the datasets which were released as part of CoNLL Shared Tasks on NER. Specifically, for English we use the data released as part of the CoNLL 2003 English NER Shared Task \cite{TjongKimSang:2003:ICS:1119176.1119195}. For Spanish and Dutch we used the data released as part of the CoNLL 2002 Shared Task \cite{TjongKimSang:2002:ICS:1118853.1118877}. The following entity tags are considered in these Shared Tasks : \text{Person}, \text{Location}, \text{Organization} and \text{Miscellaneous}. For all the three languages, the official splits are used as training, development and test files. 

Apart from these three languages we also evaluate our models on German. However, we did not have access to the German data from CoNLL (as it requires a special license). Instead we used the publicly available German NER data released by \newcite{faruqui10:_training}. This data was constructed by manually annotating the \textit{first two German Europarl session transcripts} with NER labels following the CoNLL 2003 annotation guidelines. We use the first session to create train and valid splits. Table \ref{tab:dataStat} summarizes the dataset statistics. Note that the German data is different from the English, Spanish and Dutch data which use News articles (as opposed to parliamentary proceedings). Note that the German NER data  is in IO format so, for all our experiments involving German we convert the data in other languages also to IO format. For the remaining NER experiments, data is converted to IOBES format \cite{Ratinov:2009:DCM:1596374.1596399}. 

\begin{table}[!htb]
\centering
\scriptsize
\begin{tabular}{l r r}
\toprule
\textbf{Language} & \textbf{\#Train Tokens} & \textbf{\#Test tokens} \\
\midrule
English & 204567 & 46666\\
Spanish & 264715 & 51533 \\
Dutch & 202931 & 68994 \\
German & 74907 & 20696 \\
\bottomrule
\end{tabular}
\caption{Dataset Statistics}
\label{tab:dataStat}
\end{table}

\subsection{Word Embeddings}
We used pre-trained Spectral word embeddings \cite{JMLR:v16:dhillon15a} for English, Spanish, German and Dutch. All the word embeddings are of 200 dimensions. We update these pre-trained word embeddings during training. We convert all words to lowercase before obtaining the corresponding word embedding. However, note that we preserve the case information when sending the character sequence through the CNN layer (as the case information is important for the character filters). Word embeddings for different languages lie in different feature spaces (unless we use bilingual word embeddings which are trained to reside in the same feature space). These word embeddings cannot be directly given as input to our model (as unrelated words from the 2 languages can have similar word embeddings \textit{i.e.}, similar features). We use a language dependent linear layer to map the words from the 2 languages to a common feature space in a task specific setting (common features w.r.t named entity task) and then fed these as input to the LSTM layer.

\subsection{Resource constrained setup} \label{sec:bilbowa}
In the resource constrained setup we assume that we have ample training data in one source language and only limited training data in the target language. 

In all our resource constrained experiments the LSTM parameters are always shared between the source and target language. In addition, we share one or more of the following: (i) convolutional filters (ii) space of word embeddings and (iii) decoder parameters. By sharing the space of word embeddings, we mean that instead of using individually trained monolingual Spectral embeddings for the source and target language, we use jointly trained word embeddings which project the words in a common space. We use  off-the-shelf Bilbowa algorithm \cite{icml2015_gouws15} with default settings to train these bilingual word embeddings. Bilbowa takes both monolingual and bilingual corpora as input. For bilingual corpora, we use the relevant source-target portion of Europarl corpus \cite{koehn2005epc} and \textit{Opus} \cite{SkadinsEA:LREC14}. For monolingiual corpora, we obtain short abstracts for each of the 4 languages from Dbpedia \cite{jl_2014/swj_dbpedia}.

During training, we combine the training set of the source and target languages. Specifically, we merge all sentences from the training corpus of each language and randomly shuffle them to obtain a bilingual training set. This procedure is similarly repeated for the development set.

\subsection{Hyper-parameters}
Our model contains the following hyper-parameters: (i) LSTM size, (ii) maximum width of CNN filters (iii) number of filters per width (i.e., number of filters for the same width $n$) and (iv) the learning rate. All the hyper-parameters were tuned by doing a grid search and evaluating the error on the development set. For the LSTM size we considered values from 100 to 300 in steps of 50,  for the maximum width of the CNN filters we considered values from $k$ = 4 to 9 (\textit{i.e.,} we use all filters of width 1 to $k$). We varied the number of filters per width from 10 to 30 in steps of 5 and the learning rate from 0.05 to 0.50 in steps of 0.05. 

\section{Results}
In this section we report our experimental results.

\subsection{Monolingual NER} 
The main focus of this work is to see if a resource constrained language can benefit from a resource rich language. However, before reporting results in this setup, we would like to check how well our model performs for monolingual NER (\textit{i.e.}, training and testing in the same language). Table \ref{CoNLLNER} compares our results with some very recently published state-of-the art systems. We observe that our model gives state of the art results for Dutch and English and comparable results in Spanish. This shows that a completely neural network based approach can also perform at par with approaches which use a combination of Neural Networks and CRFs \cite{yang2016multi,737}.

\begin{table}[!htb]
\centering
\footnotesize
\begin{tabular}{l p{4cm} r}
\toprule
\textbf{Language} & \textbf{System} & \textbf{F1 (\%)} \\
\midrule
\multirow{5}{*}{English} & \newcite{DBLP:journals/corr/GillickBVS15} & 86.50\\
& \newcite{yang2016multi} & \textbf{90.94} \\
& \newcite{737}  & \textbf{90.94} \\
& Our System & \textbf{90.94} \\

\midrule
\multirow{5}{*}{Spanish} & \newcite{DBLP:journals/corr/GillickBVS15} & 82.95\\
& \newcite{yang2016multi} & 84.69 \\
& \newcite{737}  & \textbf{85.75} \\
&  Our System & 84.85 \\

\midrule
\multirow{5}{*}{Dutch} 
& \newcite{DBLP:journals/corr/GillickBVS15} & 82.84 \\
& \newcite{yang2016multi}  & 85.00 \\
& \newcite{737}  & 81.74 \\
& Our System & \textbf{85.20} \\
\bottomrule
\end{tabular}
\caption{Results on Monolingual NER task}
\label{CoNLLNER}
\end{table}

\subsection{A naturally resource constrained scenario} 
We now discuss our results in the resource constrained setup. In our primary experiments, we treat German as the target language and English, Spanish and Dutch as the source language. The reason for choosing German as the target language is that the NER data available for German is indeed very small as compared to the English, Spanish and Dutch datasets (thus naturally forming a pair of resource rich (English, Dutch, Spanish) and resource poor (German) languages). We train our model jointly using the entire source (English or Dutch or Spanish) and target (German) data. We report separate results for the case when (i) the convolutional filters are shared (ii) the decoder is shared and (iii) both are shared. We compare these results with the case when we train a model using only the target (German) data. The results are summarized in Table \ref{GermanNoWodTab} (DE: German, EN: English, NL: Dutch, ES: Spanish). 

We observe that sharing of parameters between the two languages helps achieve better results compared to the monolingual setting. Sharing of decoder between English and German helps the most. On the other hand, for German and Dutch we get best results when sharing both character level filters as well as decoder parameters. For German and Spanish sharing the filters helps achieve better results.

Next, we intend to use a common word embedding space for the source and target languages where related words across the two languages have similar embeddings. The intuition here is if a source word is seen at training time but the corresponding target word (translation) is only seen at test time, the model could still be able to generalize since the embeddings of the source and target words are similar. For this, we use the jointly trained Bilbowa word embeddings as described in section \ref{sec:bilbowa}. In addition, the decoder and character filters are also shared between the two languages. These results are summarized in Table \ref{GermanBilWord}. We observe that we get larger gains when combining the source and target language data. However, the overall results are still poorer when using monolingual Spectral embeddings (as reported in Table \ref{GermanNoWodTab}). This is mainly because the monolingual corpora used for training Bilbowa word embeddings was much smaller as compared to that used for training Spectral embeddings. For example, the English Spectral embeddings were trained on a larger GigaWord corpus ($>$1 billion words) whereas the Bilbowa embeddings were trained on a smaller corpus comprising of Dbpedia abstracts (around 400 million words). Given the promising gains obtained by using these bilingual word embeddings it would be interesting to train them on larger corpora. We leave this as future work. 

\begin{table}[!htb]
\centering
\begin{subtable}[b]{0.5\textwidth}
\centering
\footnotesize
\begin{tabular}{l r r}
\toprule
\textbf{Training Data}	&\textbf{Shared} & \textbf{F1 Score}	\\
\midrule
DE	&-	&87.64	\\
\midrule
\multirow{3}{*}{DE + EN}	&Filters	& 89.05	\\
	&Decoder	& \textbf{89.89}	\\
  	&Both	& 89.08	\\
\midrule
\multirow{3}{*}{DE +
NL}	&Filters	&89.19	\\
	&Decoder	&88.48	\\
	&Both	&\textbf{89.66}	\\
\midrule
\multirow{3}{*}{DE +
ES}	&Filters	& \textbf{89.61}	\\
	&Decoder	& 89.16	\\
	&Both	&89.02	\\
\bottomrule
\end{tabular}
\caption{Using Spectral Word Embedding}
\label{GermanNoWodTab}
\end{subtable}
\begin{subtable}[b]{0.5\textwidth}
\centering
\footnotesize
\begin{tabular}{l r r}
\toprule
\textbf{Training Data}	&\textbf{Shared}	&\textbf{F1 Score}	\\
\midrule
DE &-	& 84.86	\\
DE + EN	& ALL &	\textbf{88.18} \\
\midrule
DE	&-	& 82.57	\\
DE + NL & ALL	& \textbf{87.78}	\\
\midrule
DE	&-	& 77.12	\\
DE + ES & ALL	& \textbf{85.74}	\\
\bottomrule
\end{tabular}
\caption{Using Bilingual Word Embedding}
\label{GermanBilWord}
\end{subtable}
\caption{F-Score for German Test data for Monolingual and Joint Training }
\label{fig:engNedGer}
\end{table}

\subsection{A simulated resource constrained scenario} 
\begin{figure*}[!htb]
\centering
\begin{subfigure}[b]{0.45\textwidth}
  \includegraphics[scale=0.47]{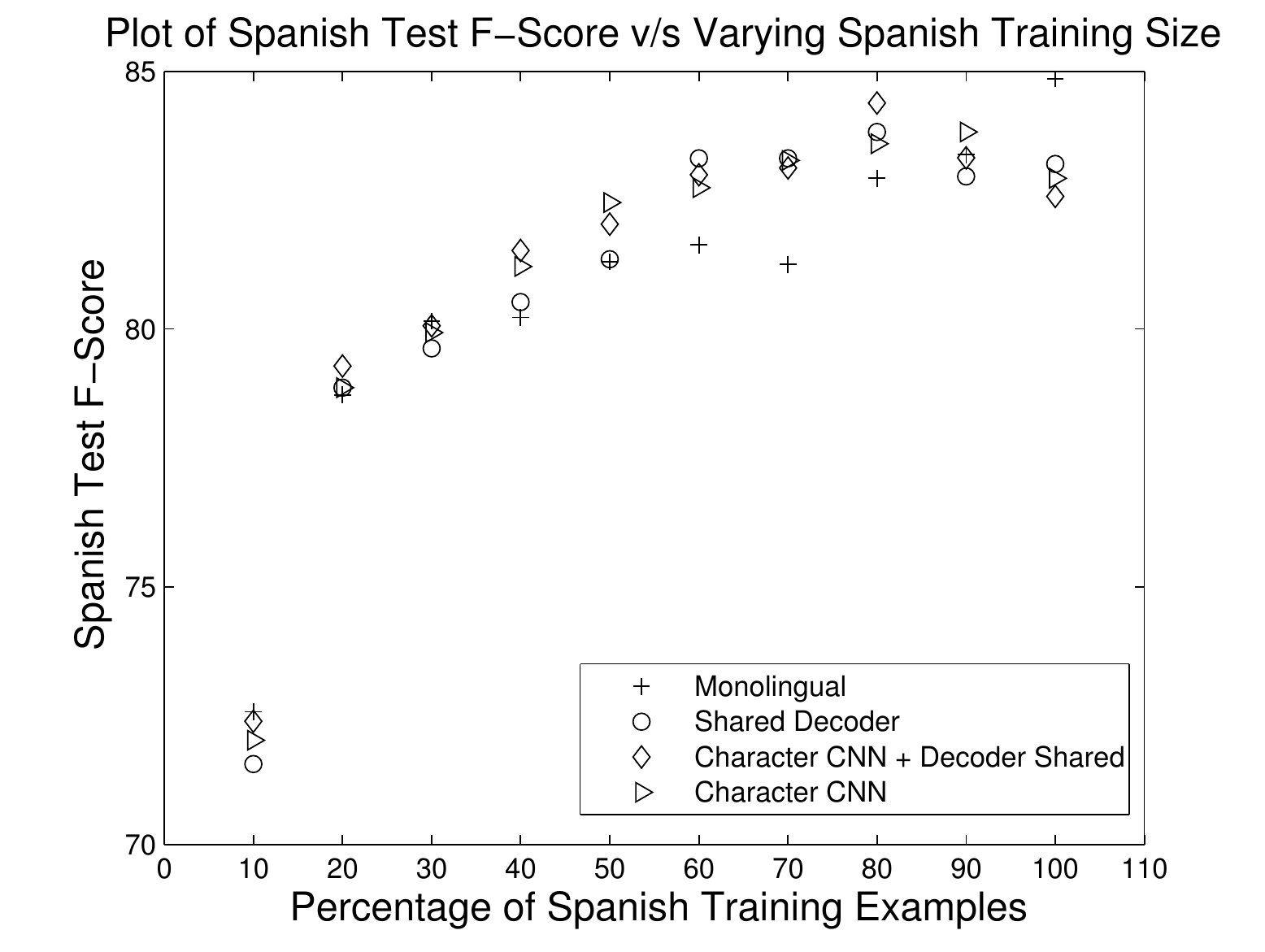}
  \caption{Sharing some parameters of the network: Using Spectral Word Embeddings}
  \label{fig:engEspShared}
\end{subfigure}
\hfill
\begin{subfigure}[b]{0.45\textwidth}
  \includegraphics[scale=0.47]{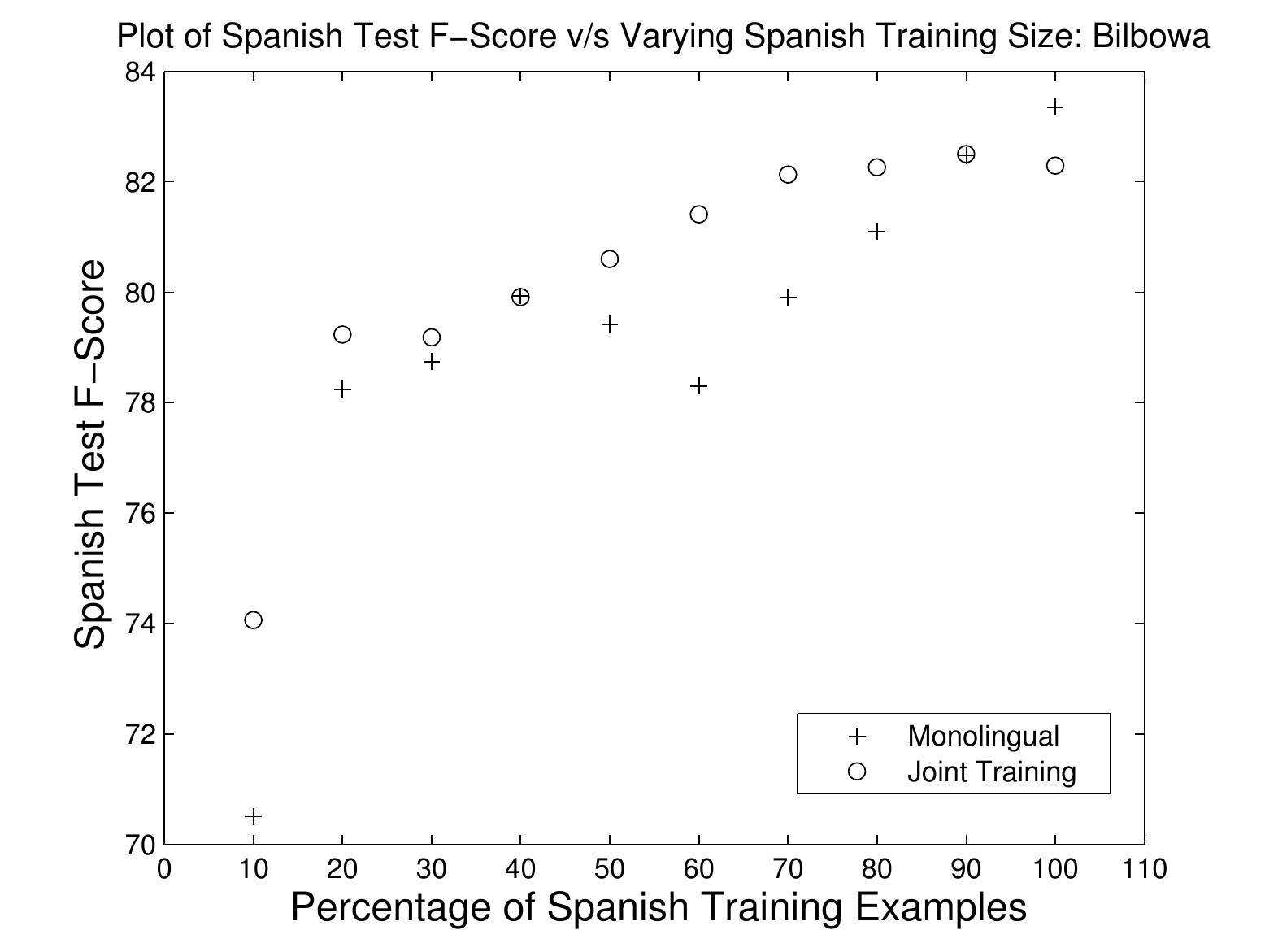}
  \caption{Sharing all parameters of the network: Using Bilingual Word Embeddings}
  \label{fig:engEspBil}
\end{subfigure}
\caption{Plot of Spanish Test F-Score for varying percentage of Spanish training data. We compared the results obtained using Monolingual training (only Spanish data) with those obtained using joint training (English and Spanish) by sharing different network parameters}
\label{fig:engEsp}
\end{figure*}

To help us analyze our model further we perform one more experiment using English as the source and Spanish as the target language. Since sufficient annotated corpora is available in Spanish, we artificially simulate a resource constrained setup by varying the amount of training data in Spanish from 10\% to 90\% in steps of 10\%. These results are summarized in Figure \ref{fig:engEspShared}. We see an improvement of around 0.73\% to 1.87\% when the amount of Spanish data is between 30\% to 80\%. The benefit of adding English data would of course taper off as more and more Spanish data is available. We hoped that the English data would be more useful when a smaller amount of Spanish data ($<$ 30\%) is available but this is not the case. We believe this happens because at lower Spanish data sizes, the English data dominates the training process which perhaps prevents the model from learning certain Spanish-specific characteristics. Finally, Figure \ref{fig:engEspBil} summarizes the results obtained when using a common word embedding space (\textit{i.e.}, using Bilbowa word embeddings) and sharing the decoder and character filters. Once again we see larger improvements but the overall results are lower than those obtained by Spectral embedding due to reasons explained above. 

\section{Analysis}
We did some error analysis to understand the effect of sharing different network parameters. Although our primary experiments were on English-German, Spanish-German and Dutch-German, we restricted our error analysis to English-Spanish since we could understand these two languages. 

\subsection{Shared Decoder}
Intuitively, sharing the decoder should allow one language to benefit from the tag sequence patterns learned from another language. Of course, this would not happen in the two languages having very different word orders (for example, English-Hindi) but this is not the case for English \& Spanish. Indeed, we observed that the Spanish model was able to benefit from certain tag sequences which were not frequently seen in the Spanish training data but were seen in the English training data. For example the tag sequence pattern \textit{ (\_O w\_LOC} is frequently confused and tagged as \textit{ (\_O w\_ORG} by the Spanish monolingual model. Here, the symbol "(" is tagged as \textit{Others} and $w$ is a place-holder for some word. However, this tag pattern was frequently observed in the English training data. For example, such patterns were observed in English Sports news articles: \textit{``Ronaldo \textbf{(\_O} \textbf{Brazil\_LOC} ) scored 2 goals in the match.''}. The joint model could benefit from this information coming from the English data and was thus able to reduce some of the errors made by the Spanish model.

\subsection{Shared Character Filters}
We observed that sharing character filters also helps in generalization by extracting language independent named entity features. For example, many location names begin with an upper-case character and end with the suffix \textit{ia} as in \textit{Australia, Austria, Columbia, India, Indonesia, Malaysia, etc.}. There were many such location named entities in the English corpus compared to the Spanish training corpus. We observed that Spanish benefited from this in the joint training setup and made fewer mistakes on such names (which it was otherwise confusing with Organization tag in the monolingual setting)

\section {Conclusion}
In this work, we focused on the problem of improving NER in a resource deprived language by using additional annotated corpora from another language. To this end, we proposed a neural network based architecture which allows sharing of various parameters between the two languages. Specifically, we share the decoder, the filters used for extracting character level features and a shared space comprising of bilingual word embeddings. Since the parameters are shared the model can be jointly trained using annotated corpora available in both languages. Our experiments involving 4 language pairs suggest that such joint training indeed improves the performance in a resource deprived language.

There are a few interesting research directions that we would like to pursue in the future. Firstly, we observed that we get much larger gains when the space of word embeddings is shared. However, due to poorer quality of the bilingual embeddings the overall results are not better as compared to the case when we use monolingual word embeddings. We would like to see if training the bilingual word embeddings on a larger corpus would help in correcting this situation. Further, currently the word embeddings are trained independently of the NER task and then fine tuned during training. It would be interesting to design a model which allows to jointly embed words and predict tags in multiple languages. Finally, in this work we used only two languages at a time. We would like to see if jointly training with multiple languages could give better results.  

\bibliography{main}
\bibliographystyle{emnlp2016}
\end{document}